\documentclass[final,3p,times]{elsarticle}

\usepackage{amssymb}
\usepackage{amsmath}
\usepackage{graphicx} 
\usepackage{lipsum}   
\usepackage{caption}  

\begin{document}

\begin{frontmatter}



\title{AutoRNet: Automatically Optimizing Heuristics for Robust Network Design via Large Language Models}


\author[label1,label2]{He Yu} 
\author[label1,label2]{Jing Liu} 
\affiliation[label1]{organization={School of Artificial Intelligence, Xidian University},
            addressline={2 South Taibai Road}, 
            city={Xi'an},
            postcode={710071}, 
            state={Shaanxi},
            country={China}}

\affiliation[label2]{organization={Guangzhou Institute of Technology, Xidian University},
            addressline={Knowledge City}, 
            city={Guangzhou},
            postcode={510555}, 
            state={Guangdong},
            country={China}}
            
\begin{abstract}
        Achieving robust networks is a challenging problem due to its NP-hard nature and the vast, complex, high-dimensional solution space. Current methods, from handcrafted feature extraction to deep learning approaches, have made certain advancements but remain rigid and complex, often requiring manual design, trial and error, and large amounts of labeled data. To deal with these problems, we propose AutoRNet, a novel framework that integrates large language models (LLMs) with evolutionary algorithms to automatically generate complete heuristics for robust network design. With the intrinsic properties of robust network structure in mind, effective network optimization strategy-based variation operations are designed to provide domain specific prompts for LLMs to help them make use of domain knowledge to generate advanced complete heuristics. Moreover, to deal with the difficulty brought by the hard constraint of maintaining degree distributions, an adaptive fitness function, which can progressively strengthening constraints to balance convergence and diversity, is designed. We evaluate the robustness of networks generated by AutoRNet’s heuristics on both sparse and dense initial scale-free networks. These solutions outperform those from current methods. AutoRNet reduces the need for manual design and large datasets, offering a more flexible and adaptive approach for generating robust network structures.
\end{abstract}



\begin{keyword}
evolutionary algorithms, large language models, complex network, network robustness, prompt engineering, deep learning


\end{keyword}

\end{frontmatter}



\section{Introduction}

Modern networked systems form the backbone of contemporary society. Understanding the robustness of these networks is crucial for ensuring stability and reliability \cite{watts1998, barabasi1999, schneider2011}. Improving network robustness to prevent catastrophic disruptions is inherently challenging due to its NP-hard nature, and related research has seen significant advancements. Key contributions include approximate theoretical models \cite{tra2013}, which simplify the complex interactions within networks to provide insights into enhancing resilience. Optimization algorithms, employing approaches such as simulated annealing (SA) \cite{buesser2011}, genetic algorithms (GAs) \cite{zhou2014}, and greedy approaches \cite{zeng2012}, offer near-optimal solutions in a reasonable time frame. In addition, machine learning techniques, particularly deep reinforcement learning\cite{DBLP:journals/corr/abs-1906-09959}, have been used to dynamically adapt and improve network configurations. However, these methods highly depend on manual work, expert knowledge, training data, and trial-and-error processes.

The emergence of coding-oriented Large Language Models (LLMs)\cite{naveed2023LLMs} has garnered significant attention for their potential in addressing combinatorial optimization problems \cite{silva2022, ahmadi2023}. FunSearch \cite{romera2024} combines a pre-trained LLM with evolutionary algorithms (EAs) to evolve initial low-scoring programs into high-scoring ones. Evolution of Heuristics (EoH) \cite{liu2024} co-evolves both heuristic descriptions and their corresponding code implementations. However, both FunSearch and EoH depend primarily on existing optimization algorithms, only scoring or weighting mechanisms influencing the search algorithm behavior are designed by LLMs to guide the related operations, without creating a new algorithm, hereby limiting their applicability to complex domain-specific problems, such as network robustness. \emph{There is a need to generate whole algorithms directly rather than merely modifying data through weighting and scoring.}

In this paper, we address these issues and make a significant step by proposing AutoRNet, an integrated framework that combines the contextual intelligence of LLMs with the adaptive optimization capabilities of EAs. Specifically, we design Network Optimization Strategy(NOS)-based variation operations tailored for complex network problems, which can create domain-specific prompts for LLMs, generating a variety of heuristics suitable for different network-related challenges. Moreover, we design an Adaptive Fitness Function (AFF) to evaluate these heuristics, progressively tightening constraints to balance the convergence and diversity, thereby discovering superior heuristics. AutoRNet can automatically make use of special network characteristics to design effective operations. The major contributions of this paper are summarized as follows:
\begin{itemize}
\item A hybrid framework is developed where EAs and LLMs iteratively collaborate to generate and refine heuristics for network robustness.
\item NOS-based variation operations are designed, which can generate problem-specific prompts to guide LLMs in divergent thinking. These NOSs are equally applicable to other network-related challenges.
\item AFF tailored to network issues is designed, to transform hard constraints into soft ones, enhancing the diversity and resilience of heuristics.
\item Solutions generated by AutoRNet's heuristics are evaluated across eight scale-free networks with varying sizes and densities and a real-world network, demonstrating that they outperform current methods.
\end{itemize}

The remainder of this paper is organized as follows. Sections 2 and 3 introduce the related work and network robustness measures, respectively. Section 4  introduces AutoRNet in details. Section 5 presents the experiments. Finally, Section 6 concludes the paper with a summary of our findings.

\section{Related Work}

\subsection{Related Work on Improving Network Robustness}

    Traditional methods for improving network robustness focus on improving local redundancy and connectivity, such as maximizing \emph{clustering coefficient} to form tightly knit groups of nodes. These methods also leverage high-order network structures like triangles and quadrilaterals to identify and protect \emph{critical nodes} and \emph{edges} to withstand targeted attacks. Techniques like \emph{greedy} and \emph{local search algorithms} systematically improve network robustness. Fortunato et al.\cite{fortunato2006} proposed a greedy algorithm that improves the robustness of social networks by forming tightly-knit communities. Zhang et al.\cite{zhang2009} proposed a lazy-greedy algorithm that enhances the robustness of transportation networks by protecting the most critical nodes from failure. Hau Chan et al.\cite{Chan2016OptimizingNR} introduced a local search algorithm by iteratively improving the subgraph structure to improve the robustness of the network.
    
    Metaheuristic algorithms, including \emph{Genetic Algorithms(GAs)}\cite{holland1975adaptation}, and \emph{Simulated Annealing(SA)}\cite{kirkpatrick1983optimization}, have also been extensively employed to solve network robustness problems. Pizzuti et al.\cite{dai2008} proposed using GAs to enhance the robustness of complex networks by simulating the process of natural selection to evolve network configurations over generations, improving resilience against failures and attacks. Zhou et al.\cite{zhou2014} proposed using Memetic Algorithms, which combine global and local search strategies, to enhance the robustness of scale-free networks by integrating global and local search operators. Buesser et al.\cite{buesser2011} proposed the use of SA to optimize the robustness of scale-free networks by rewiring the network edges, thus improving the resilience of the network to fragmentation and intentional damage. Pizzuti et al.\cite{pizzuti2008} introduced RobGA, a GA designed to enhance network robustness by adding edges in a way that minimizes disruption risk.

    However, these methods often require manual design and multiple trial-and-error processes, making them inefficient and time-consuming. The need for extensive experimentation and tuning reduces their practicality for large-scale and dynamic network environments, highlighting the need for more automated and adaptive approaches to network robustness optimization.

    Deep learning, particularly \emph{Graph Neural Networks (GNNs)}, has emerged as a powerful approach to address network robustness problems. Tang et al.\cite{tang2019} explored methods to improve the robustness of GNNs against poisoning attacks by leveraging clean graphs from similar domains. This approach helps the GNNs detect adversarial edges more effectively, enhancing their resistance to attacks. Wang et al.\cite{wang2020} developed certifiably robust GNNs to defend against attacks that perturb the graph structure by adding or deleting edges. Their approach ensures that the GNNs' performance remains stable even under adversarial conditions.

    Although GNN-based methods for network robustness are powerful, they face several challenges. These models require large amounts of labeled data for training, which can be difficult to obtain in practice. The high computational demands for training and inference can also be a limitation, especially in real-time applications. Moreover, GNNs can still be vulnerable to sophisticated adversarial attacks, where small, carefully crafted perturbations in the input data can lead to incorrect classifications.

\subsection{The Application of LLMs in Combinatorial Optimization}

LLMs have shown a significant impact in addressing various problems. One prevalent approach involves \emph{engineering in-context learning prompts (EILP)}\cite{schulhoff2024promptreportsystematicsurvey} using techniques such as \emph{zero-shot, few-shot, and chain-of-thought (CoT) prompting}\cite{kojima2022zeroshot, brown2020gpt3, wei2022cot}. Enhancing EILP with \emph{fine-tuning} further improves response accuracy by adapting LLMs to specific datasets or tasks. Furthermore, combining EILP with \emph{ensemble learning} techniques increases robustness and consistency by aggregating multiple model predictions. Integration with \emph{Retrieval-Augmented Generation (RAG) }\cite{gao2023retrieval} uses external knowledge sources, allowing LLMs to generate more accurate and informed contextual responses.

Many researchers have also focused on solving combinatorial optimization problems using LLMs. Shengcai Liu et al.\cite{liu2024largeasEO} presented the first investigation into LLMs as evolutionary combination optimizers for solving the Travelling Salesman Problem (TSP). FunSearch\cite{romera2024} leverages LLMs to generate and optimize mathematical functions, discovering new constructions and improving existing solutions by iteratively refining function constructions. EoH\cite{liu2024} facilitates the simultaneous evolution of concept description and code implementations, emulating the process by which humans develop heuristics, to achieve efficient automatic heuristic design. These approaches have demonstrated competitive performance compared to traditional heuristics in finding high-quality solutions.

All of these methods are based on existing algorithms, scoring or weighting mechanisms are used to modify the original data (e.g., bin capacities, city distances), influencing the search algorithm behavior but without creating a new algorithm. \emph{There is a need to generate whole algorithms directly rather than merely modifying data through weighting and scoring.} These approaches restrict their optimization improvements to relatively simple combinatorial problems and \emph{cannot extend their applicability to more complex or domain-specific optimization scenarios. }

\section{Network Robustness Measures}

A network is often represented as a graph $G = (V, E)$, where $V$ denotes the set of nodes and  $E$  represents the set of edges. To measure network robustness, several methodologies have been developed. \cite{schneider2011} introduced a measure $R$ to assess network robustness by evaluating the size of the largest connected component during sequential node attacks,
\begin{equation}
R = \frac{1}{N} \sum_{q=1}^{N} s(q)
\end{equation}
where \( N\) is the number of nodes in the network and \( s(q) \)  is the fraction of nodes in the largest connected cluster after removing \(q\) nodes. The normalization factor \(1/N \) ensures that the robustness of networks with different sizes can be compared. The range of \( R \) values is between \(1/N \) and 0.5.

Complex networks can exhibit different topological structures, with random and scale-free networks being two of the most studied types.  Random networks, characterized by a homogeneous degree distribution, are robust to targeted attacks on high-degree nodes. In contrast, scale-free networks, with a power-law degree distribution, exhibit exceptional robustness against random failures due to the low probability of removing a hub but are extremely susceptible to targeted attacks that focus on their few high-degree hubs. These networks display distinct characteristics that significantly impact their robustness and vulnerability to failures or attacks.

In this paper, we adopt $R$ to evaluate network robustness, specifically focusing on targeted attacks through the sequential removal of the highest-degree nodes. We employ scale-free networks for both training and testing graph sets to accurately reflect the structure of many real-world networks.

\section{AutoRNet}

\subsection{The Framework of AutoRNet}

\begin{figure*}[t]
\centering
\includegraphics[width=1\textwidth]{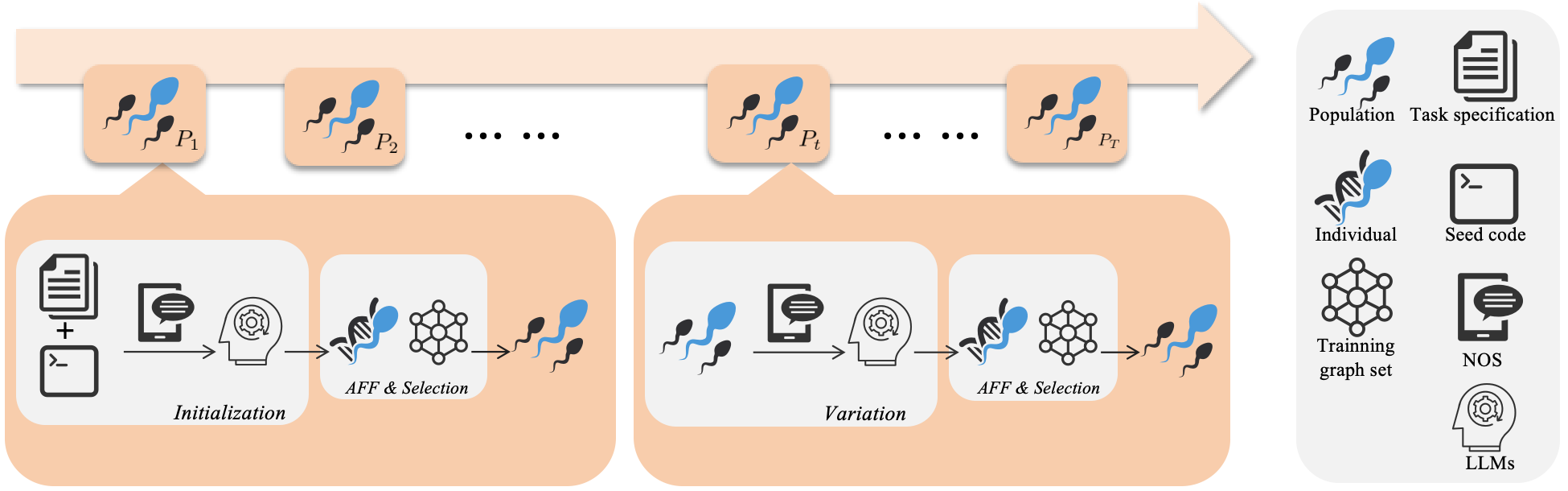} 
    \caption{A schematic illustration of AutoRNet}
    \label{fig:AutoRNet_Framework}
\end{figure*}

AutoRNet uses EAs to search heuristics and maintains a population of \emph{popsize} individuals, denoted as \(P = \{ h_{1}, h_{2}, \ldots, h_{popsize} \}\). Each individual \(h_i\), \(i=1, 2, \ldots, popsize\), includes a heuristic. There are \(T\) generations in total, with \(P_{t} = \{ h_{t,1}, h_{t,2}, \ldots, h_{t,popsize} \}\) representing the population at generation \(t\). AutoRNet evolves the population generation by generation, and the whole framework is summarized in Algorithm 1.

First, \emph{InitializePopulation}() initializes the initial population \(P_1\) of $popsize$ individuals using a \emph{task specification} and prompt interaction with the LLM. Then, based on the training graph set \(\mathcal{G}\), \emph{AFF}() uses the adaptive fitness function, introduced in the following text, to calculate the fitness of each individual in \(P_1\). Next, the population is evolved \(T\) generations, and in each generation, \emph{NOS\_Variation}() first conducts the NOS-based variation operations designed in the following text on the current population, obtaining the offspring population. Then, \emph{SelectNextPopulation}() uses the roulette wheel selection according to the fitness to select $popsize$ individuals from \(P_t\) and \(P_{offspring}\) together to form the population for the next generation.  Figure \ref{fig:AutoRNet_Framework} schematically illustrates the framework of AutoRNet.

\begin{figure}[t]
\centering
\includegraphics[width=0.5\columnwidth]{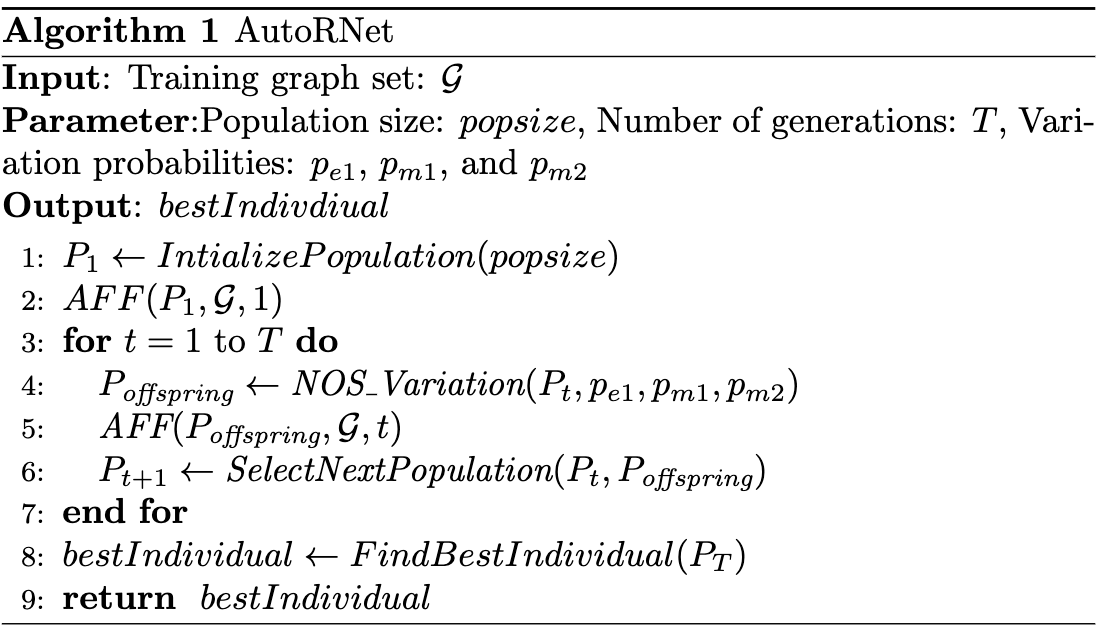}
\end{figure}

\subsection{Individual Encoding and Population Initialization}

\begin{minipage}[c]{0.49\textwidth}
\centering
\includegraphics[width=0.9\columnwidth]{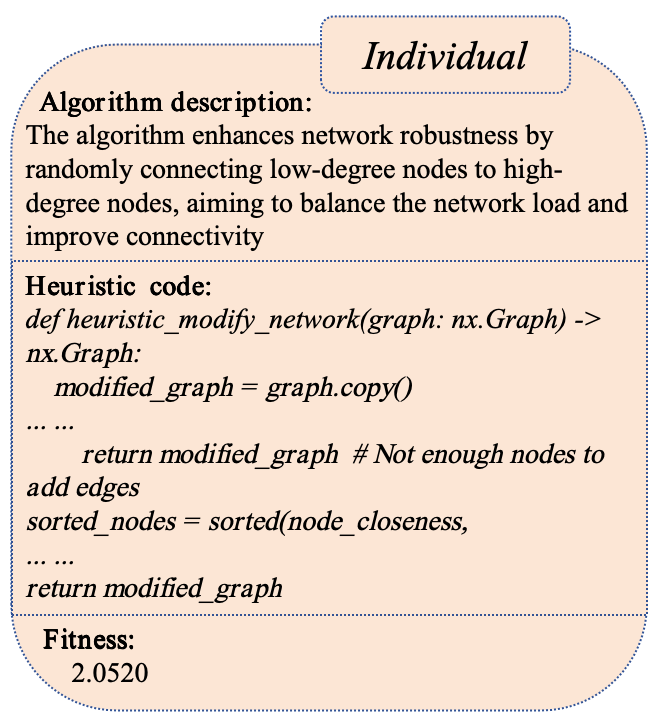}
    \captionof{figure}{Individual structure schematic.}
    \label{fig:individual_structure}
    \end{minipage}
    \hfill
    \begin{minipage}[c]{0.49\textwidth}
\centering
\includegraphics[width=1\columnwidth]{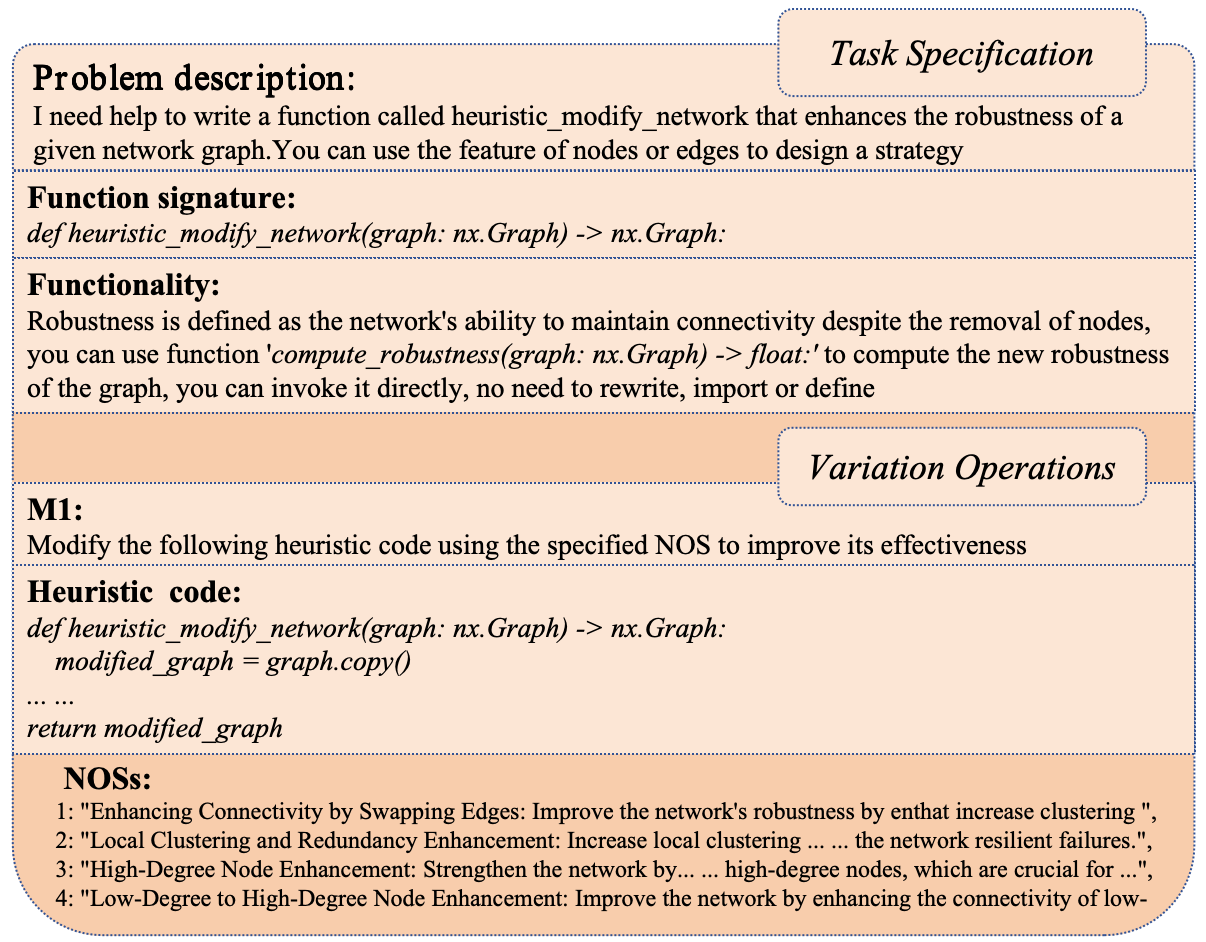}
    \captionof{figure}{Variation Operation prompt for M1}
    \label{fig: Variation Operation prompt}
\end{minipage}
\\
\\

 The individual's heuristic codes are generated by LLMs. Each individual is structured as follows, which is schematically illustrated in Figure \ref{fig:individual_structure}: 
\begin{itemize}
\item \textbf{Algorithm Description}: A natural language explanation of the heuristic’s objective, providing context and understanding of what the heuristic aims to achieve.
\item \textbf{Heuristic Code}: The actual implementation of the heuristic, detailing the logic and operations needed to improve network robustness.
\item \textbf{Fitness}: The fitness of the heuristic evaluated by \emph{AFF}.
\end{itemize}

AutoRNet initializes its heuristic population by providing LLMs with a detailed \emph{task specification} that includes \emph{a description of the problem, function signature, and functionality of the heuristic}. The LLM generates initial heuristics by creating and implementing them as Python code, with the process repeated multiple times to ensure diversity. Optionally, \emph{heuristic seed codes}, either simple or expert-designed, are included to guide the LLM towards effective methods. Existing hand-crafted heuristics can be also incorporated to enhance the initial population with additional domain knowledge. The detailed population initialization prompts are provided in the \textbf{Appendix}.

\subsection{Adaptive Fitness Function}

In the EA assisting LLMs for heuristic code generation, individuals are encoded as methods rather than solutions, and the search is based on LLMs to generate \emph{new} or \emph{similar} methods. \emph{It does not inherently provide a mechanism to measure the similarity between methods, making it difficult to navigate the method space.} Maintaining a consistent degree distribution is a constraint in optimizing network robustness. Enforcing this constraint strictly from the outset can be overly restrictive, leading to a large number of invalid individuals, impeding the evolutionary process.

To overcome these issues, we design an \emph{adaptive fitness function (AFF)}, which can dynamically adjust the degree distribution constraint throughout the evolutionary process. The AFF initially relaxes this constraint, facilitating broader exploration of the method space, and then progressively tightens this constraint as the evolution proceeds. By doing so, \emph{AFF improves the ability of AutoRNet to explore various heuristics and avoid dropping into local optima.}

AFF uses a training graph set to evaluate the performance of heuristics obtained by each individual adaptively with the evolution process. The training graph set  \(\mathcal{G} = \{G_1, G_2,\ldots, G_M\}\) consists of $BA$ scale-free graphs. To calculate the fitness of \(h_{t,i}\), the heuristic  \(H_{t,i}\) of  \(h_{t,i}\) is first used to optimize  \(\mathcal{G}\) to \(\tilde{\mathcal{G}}_{t,i}\)
\begin{equation}
    H_{t,i}(\mathcal{G}) = \tilde{\mathcal{G}}_{t,i}
    \label{eq:heuristic_fun}
\end{equation}
where \(\tilde{\mathcal{G}}_{t,i} = \{\tilde{G}_{t,i,1}, \ldots, \tilde{G}_{t,i,M}\}\). Here, \(\tilde{G}_{t,i,j}\) denotes the \(j\)-th graph obtained by optimizing \(G_j\) using \(H_{t,i}\). The AFF evaluates the fitness of  \(h_{t,i}\) as follows:

\begin{equation}
\label{eq:AFF}
f(\tilde{\mathcal{G}}_{t,i}, t) = \sum_{j=1}^{M} \left( R(\tilde{G}_{t, i, j}) \cdot \left( 2 -  w(t) \cdot Y(\tilde{G}_{t, i, j})\right) \right)
\end{equation}
where $Y(\tilde{G}_{t,i,j})$ quantifies the deviation of the optimized graph  $\tilde{G}_{t,i,j}$ from the original graph $G_j$, defined as: 
\begin{equation}
Y(\tilde{G}_{t,i,j}) =  \frac{D_{\text{diff}}(\tilde{G}_{t, i, j})}{D_{\text{max}}(\tilde{G}_{t, i, j})} + \frac{E_{\text{diff}}(\tilde{G}_{t, i, j})}{E_{\text{max}}(\tilde{G}_{t, i, j})} 
\end{equation}
where \( D_{\text{diff}}(\tilde{G}_{t, i, j}) \) is the difference in degree distribution between original graph \( G_j \) and optimized graph \( \tilde{G}_{t, i, j} \). 
\begin{equation}
D_{\text{diff}}(\tilde{G}_{t, i, j}) = \frac{1}{N} \sum_{k=1}^{N} |d_k(G_j) - d_k(\tilde{G}_{t, i, j})|
\end{equation}
The maximum possible deviation in degree distribution \( D_{\text{max}}(\tilde{G}_{t, i, j}) \) is calculated as follows:

\begin{equation}
D_{\text{max}}(\tilde{G}_{t, i, j}) =\max_{k \in \{1,2,...,N\}} |d_k(G_j) - d_k(\tilde{G}_{t, i, j})|
\end{equation}
The normalized degree distribution difference \(\frac{D_{\text{diff}}(\tilde{G}_{t, i, j})}{D_{\text{max}}(\tilde{G}_{t, i, j})}\) has a value range between 0 and 1.

Similarly, \( E_{\text{diff}}(\tilde{G}_{t, i, j}) \) represents the difference in edge count between the original graph \( G_j \) and the optimized graph \( \tilde{G}_{t, i, j} \). \( E_{\text{diff}}(\tilde{G}_{t, i, j}) \) is defined as:
\begin{equation}
E_{\text{diff}}(\tilde{G}_{t, i, j}) = |E(G_j) - E(\tilde{G}_{t, i, j})|
\end{equation}
where \( E(G_j) \) and \( E(\tilde{G}_{t, i, j}) \) represent the edge count in the original and optimized graphs, respectively. 

\( E_{\text{max}}(\tilde{G}_{t, i, j}) \) is the maximum possible deviation in edge count, calculated as follows:
\begin{equation}
E_{\text{max}}(\tilde{G}_{t, i, j}) = {\max(E(G_j),  E(\tilde{G}_{t, i, j}))}
\end{equation}

The normalized edge number difference \(\frac{E_{\text{diff}}(\tilde{G}_{t, i, j})}{E_{\text{max}}(\tilde{G}_{t, i, j})}\) has a value range between 0 and 1.

The weight function \( w(t) \) increases the penalty on structural deviations as generations progress, defined as:

\begin{equation}
\label{eq:w(t)}
w(t) = \left(\frac{t}{T}\right)^p
\end{equation}
where \( T \) is the total number of generations and \( p \) is a parameter controlling the rate of increase. Typically, \( p \) is chosen in the range \( 0.5 \leq p \leq 2 \). 

The key idea of $f(\tilde{\mathcal{G}}_{t,i}, t)$ in Equation \ref{eq:AFF} is: If the optimized \(\tilde{G}_{t, i, j}\) satisfies the consistency constraints for both edge count and degree distribution (that is, both \( D_{\text{diff}} \) and \( E_{\text{diff}} \) are $0$), it will receive a reward double the base score (\( 2 \times R(\tilde{G}_{t, i, j}) \)). If these constraints are not met, the penalty for structural deviations increases over successive generations. Consequently, the fitness value ranges from 0 to \( 2 \times R(\tilde{G}_{t, i, j}) \).

\subsection{\textbf{Network Optimization Strategy-based Variation Operations}}

The heuristic method searching space in the EA assisting LLMs for heuristic code generation is vast, complex, and high-dimensional. In simpler problem domains, such as those addressed by EoH and FunSearch, this space is reduced by limiting function codes to straightforward tasks like weighting or scoring data. For complex problems like network robustness, simplifying the problem is not feasible due to the intricate and domain-specific nature of tasks. Through experiments we find, to deal with the complex problem like network robustness, by providing just general prompts without domain knowledge, current LLMs can only design simple operations on nodes or links, and lack of the ability to make deep use of domain knowledge to design advanced operations. Therefore, it is important to design mechanism which can provide LLMs domain knowledge effectively to further release LLMs’ ability in design optimization methods for complex problems.

 To cure this problem, we design \emph{Network Optimization Strategies (NOSs)} with the intrinsic properties of networks in mind. Networks have \emph{features} such as degree distribution, path characteristics, clustering coefficient, centrality measures, and community structure. Based on these features, \emph{strategies} such as high-degree node priority, shortest path optimization, and betweenness centrality priority can be designed. Guided by these \emph{strategies}, \emph{actions} such as adding edges, rewiring edges, and swapping edges are then taken. \emph{This combination of features, strategies, and actions forms the NOS}. Detailed information is provided in the \textbf{Appendix}.

Based on NOSs, variation operations are designed to generate offspring heuristics by integrating \emph{NOSs} into prompts to guide the LLMs with domain knowledge. Three types of variation operations, namely \textbf{E1}, \textbf{M1}, and \textbf{M2}, are designed. E1 and M1 form the prompt for LLMs by integrating randomly select 12 NOSs with the general purpose prompts, and M2 just use the general prompts.

\textbf{E1 (Exploration with NOS Integration)}: By providing 2 parent individuals, E1 prompts the LLM to create entirely new heuristics with randomly selecting 12 NOSs, ensuring that the generated offspring are diverse and innovative. E1 help AutoRNet escape local optima by introducing new strategies and methods into the population.

\textbf{M1 (Guided Modification with NOS)}: M1 prompts the LLM to refine existing heuristics by incorporating NOSs leading to targeted improvements and optimizations. This type of variation operation provides guided local search, leveraging domain-specific knowledge to enhance the effectiveness of the current heuristic.

\textbf{M2 (Local Adjustment)}: M2 prompts the LLM to make minor adjustments to existing heuristics, focusing on small-scale improvements and refinements. This type of variation operation is a pure local search, making incremental adjustments to optimize the heuristic’s performance.

In each generation, $p_{e1} \times popsize$, $p_{m1} \times popsize$, $p_{m2} \times popsize$ individuals are selected from the current population using the tournament selection to conduct E1, M1, M2, respectively. In this way, prompts of E1, M1, and M2 can be provided to the LLM server simultaneously. Figure \ref{fig: Variation Operation prompt} illustrates the prompt schematic for M1, and these for E1 and M2 are given in the \textbf{Appendix}. By utilizing these three types of variation operations, AutoRNet effectively balances the need for innovation and refinement, ensuring robust and efficient network optimization.

\section{Experiments}
AutoRNet generates heuristic methods and we analyze these methods to illustrate the design capabilities of AutoRNet. Simultaneously, we select three existing algorithms as baselines: the Hill Climbing Algorithm (HC)\cite{paterson2018}, the Simulated Annealing Algorithm (SA)\cite{buesser2011}, and the Smart Rewiring Algorithm (SR)\cite{louzada2013}. $R$ is used to evaluate the robustness of networks in the test graph set after optimization by all algorithms. By comparing their network robustness, we assess the effectiveness of the methods generated by AutoRNet.

\subsection{Experimental Settings}

The training graph set $\mathcal{G}$ consists of BA scale-free networks in two different sizes: 50 and 100 nodes. For each network size, the number of initial nodes \(M_0\) varies from 2 to 5. For each combination of network size and \(M_0\), we create three instances, resulting in a total of \( M = 2 \times 4 \times 3 = 24\) training graphs. The test graph set consists of three types of networks: sparse BA networks with 100, 200, 300, and 500 nodes, $N_0 = 3$, $M_0 = 2$;  BA networks with 100 nodes, $N_0 = 6$, $M_0$ ranging from 2 to 5; and a real world EU power grid network\cite{zhou2005} with 1,494 nodes and 2,066 edges.

Each heuristic method performs a series of network modifications, such as edge addition, relocation, and swapping. After each modification, $R$ is used to evaluate the network's robustness. If $R$ improves, the modification is accepted; otherwise, it is rolled back. Consequently, the $R$ function is called multiple times to guide the optimization process. The total number of such evaluations for each heuristic is defined as \(max\_attempts\). For the training phase, \(max\_attempts\) is set to 100. For the testing phase, \(max\_attempts\) is set to $3 \times 10^4$ for the BA networks and $5 \times 10^4$ for the EU power grid network, which is the same with those of the three baseline algorithms used.

The parameters of AutoRNet were set as follows: \emph{popsize}, \emph{T}, $p_{e1}$, $p_{m1}$, and $p_{m2}$ are set to 10, 50, 0.8, 0.1, and 0.1, respectively. The GPT-4 Turbo model was used with a temperature setting of 1. \( p \) in Equation \ref{eq:w(t)} was set to 1.5. The experiments involving networks with 100 to 300 nodes were conducted over 100 independent runs, for networks with 500 nodes,  over 30 independent runs, and for the EU power grid network, over 10 independent runs.

\subsection{Evaluation Results}

\subsubsection{Design Capabilities of AutoRNet}
After running AutoRNet for 50 generations, we selected three highly valuable heuristics based on their fitness values, named as \emph{Heuristic-v1, Heuristic-v2} and \emph{Heuristic-v3}, which significantly improved the network robustness of the test graph set. \emph{Heuristic-v1}  leverages network features of critical nodes and similar nodes, \emph{Heuristic-v2} utilizes node connectivity and degree distribution, and \emph{Heuristic-v3} optimizes local topology by manipulating edges among neighbors. All these three heuristics maintain the same number of edges, with \emph{Heuristic-v1} also preserving the degree distribution. This demonstrates AutoRNet’s ability to design complete algorithms that effectively utilize network features.

\textbf{Heuristic-v1}, shown in Algorithm 2, employs an advanced strategy that combines edge swapping with SA while preserving the degree distribution. This heuristic identifies critical nodes(those with the highest degrees) and pairs similar nodes based on their degree deviation. It dynamically adjusts the \textit{max\_diff} parameter(Step 4), which controls the tolerance for degree deviation in pairing similar nodes(Steps 22-34). The algorithm iteratively swaps edges between the neighbors of these paired nodes(Steps 18-19), evaluating new configurations based on robustness improvements or probabilistic acceptance criteria derived from simulated annealing(Step 22). This sophisticated approach ensures a balance between exploration and exploitation in the search space. Without NOSs, it would be impossible to evolve such “intelligent” heuristics that smartly leverage network features for robust network design.

\begin{figure*}[t]
\centering
\includegraphics[width=1\textwidth]{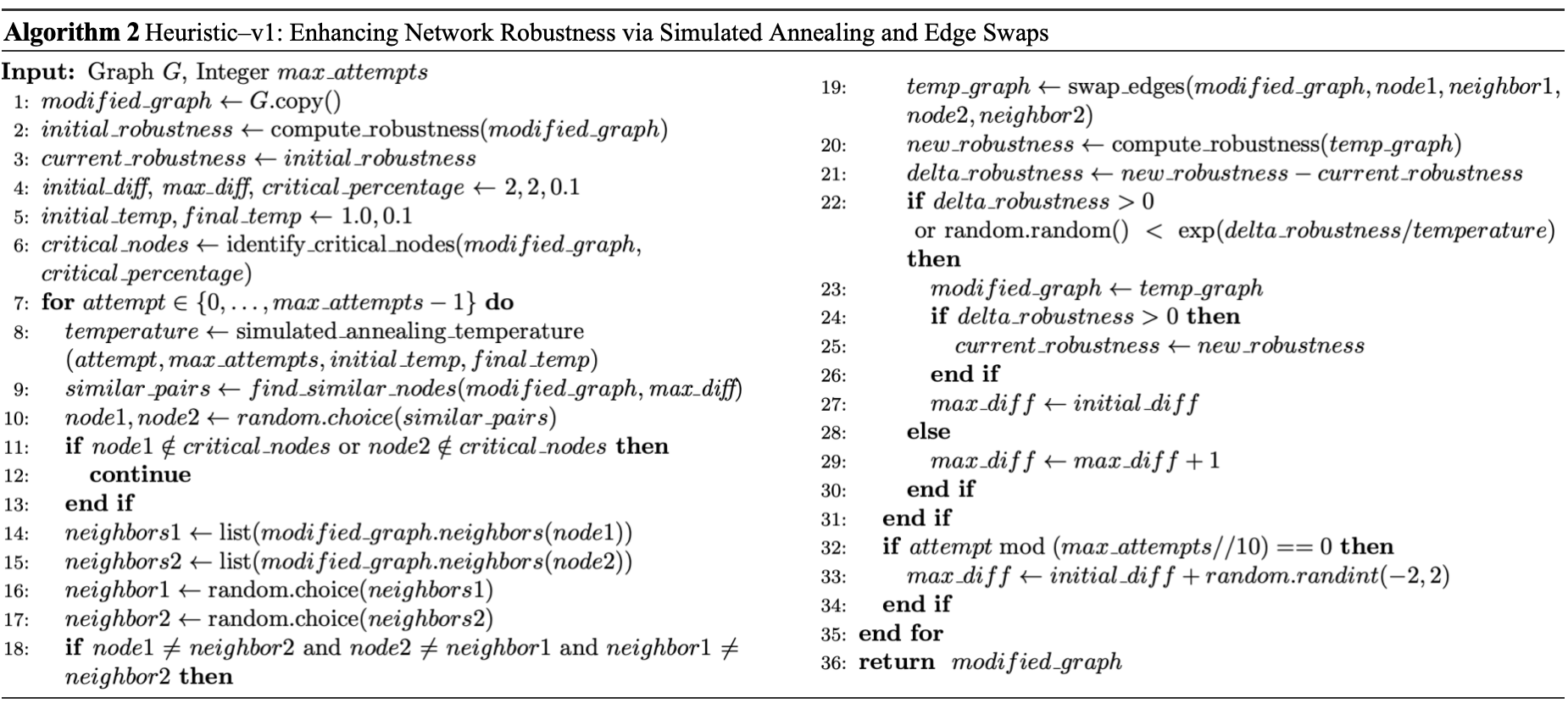} 
\end{figure*}

\emph{Heuristic–v2 and Heuristic–v3} optimize the network through edge relocation, which, while slightly altering the degree distribution, significantly improves the robustness of the networks. Importantly, edge-relocation actions incur the same real-world costs as edge-swapping. \emph{This shows that AutoRNet is not constrained by theoretical conditions, but instead explores a variety of methods, making it more suitable for solving real-world problems.}
\begin{minipage}[c]{0.5\textwidth} 
\includegraphics[width=0.95\columnwidth]{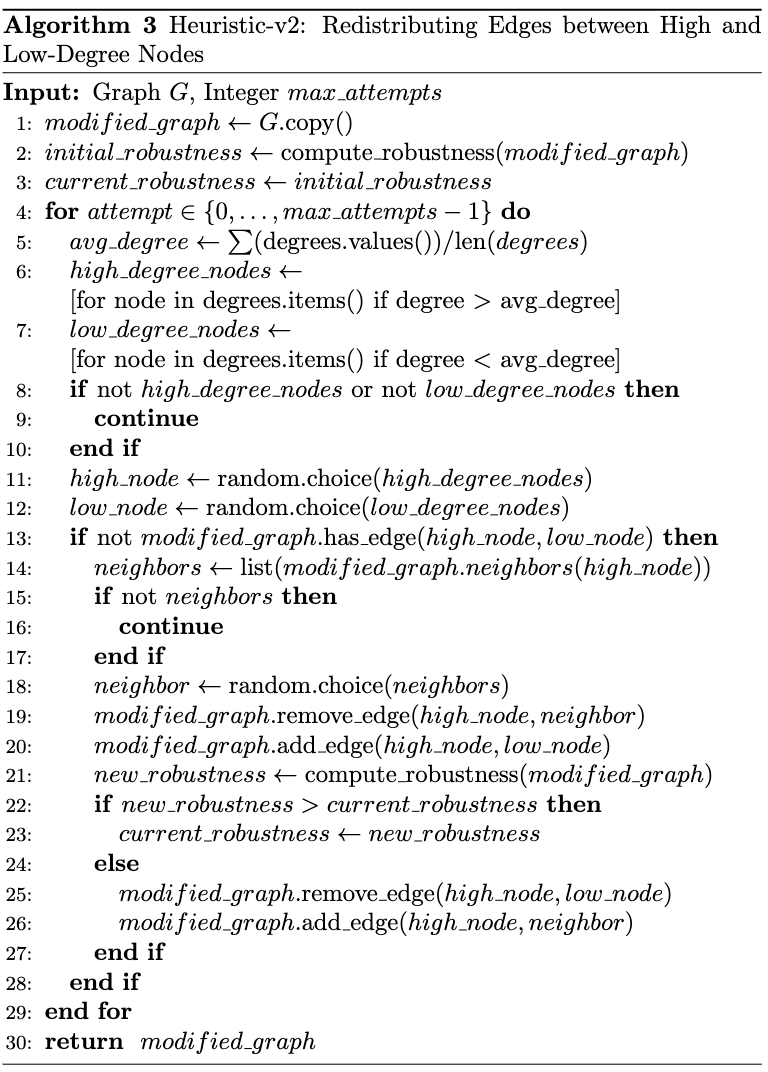}
\end{minipage}
\hfill
\begin{minipage}[c]{0.5\textwidth}  
\centering
\includegraphics[width=0.95\columnwidth]{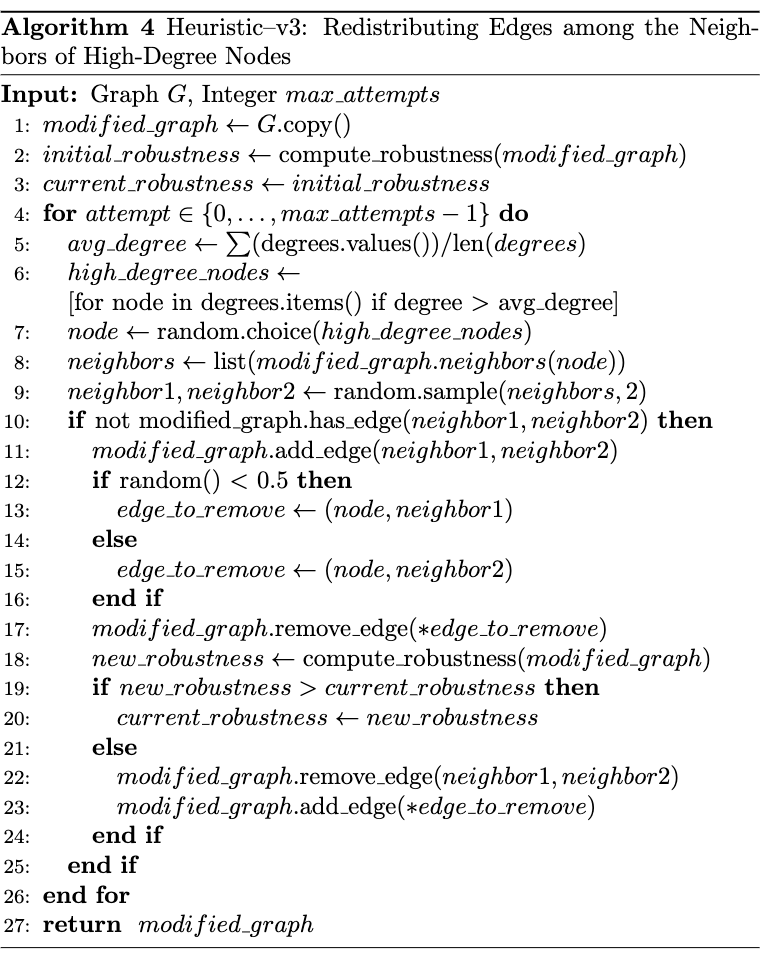}
\end{minipage}

\hfill
\\

\textbf{Heuristic-v2}, shown in Algorithm 3, improves network robustness by redistributing edges between high-degree and low-degree nodes. The heuristic involves adding an edge between a high-degree node and a low-degree node while removing an edge from the high-degree node.

\textbf{Heuristic-v3}, shown in Algorithm 4, enhances robustness by redistributing edges among the neighbors of high-degree nodes. This method identifies high-degree nodes and strategically adds edges between their neighbors, maintaining the overall edge count but improving resilience to failures.

The capabilities of AutoRNet to design heuristics that either strictly adhere to or flexibly navigate \emph{AFF} constraints highlights its versatility. AutoRNet not only matches the complexity of manually designed algorithms but also explores new strategies that yield better performance at the same practical cost. This leads to broader considerations about the potential of automated heuristic design in optimizing complex networks of the real world.

\subsubsection{Performance Comparison of Algorithms}

The robustness results over all test graphs, summarized in Tables \ref{tab:table_results_different_nodes}-\ref{tab:table_results_real_world}, demonstrate the effectiveness of the heuristics designed by AutoRNet.

\textbf{Heuristic-v1}’s performance is not as strong as those of \emph{Heuristic-v2} and \emph{Heuristic-v3}, but it is comparable to those of the three manually designed algorithms, often surpassing their in certain scenarios. In Table \ref{tab:table_results_real_world}, \emph{Heuristic-v1} performs comparably well on the EU Power Grid Network, maintaining robustness with an average of 0.205911 and a low variance, often surpassing the baseline algorithms in stability.

Specifically, \emph{Heuristic-v1} maintains stable performance with low variance across all network sizes and densities. For instance, in Table \ref{tab:table_results_different_nodes}, \emph{Heuristic-v1} surpasses the baseline algorithms in the 100 nodes network scenario by achieving a lower variance 0.000069. In Table \ref{tab:table_results_different_densities}, for edge density \(M_0 = 2\), \emph{Heuristic-v1} shows better average robustness 0.262601 with lower variance than the baseline algorithms.

\textbf{Heuristic-v2} consistently achieves the best performance across most test cases, demonstrating its superior robustness enhancement capabilities. However, it does not always outperform other algorithms in every scenario. For example, in Table \ref{tab:table_results_different_densities}, for denser networks with \(M_0 = 4\) and \(M_0 = 5\), \emph{Heuristic-v2} does not achieve the highest robustness, being outperformed by SA and \emph{Heuristic-v3}. Nonetheless, \emph{Heuristic-v2} remains one of the top-performing algorithms overall, excelling in the majority of scenarios.

\textbf{Heuristic-v3}, although not as strong as \emph{Heuristic-v2}, consistently outperforms the baseline algorithms. For example, in Table \ref{tab:table_results_different_nodes}, \emph{Heuristic-v3} shows strong performance in larger networks, achieving an average robustness of 0.307550 for 200 nodes and 0.307023 for 300 nodes, outperforming all baseline algorithms. In Table \ref{tab:table_results_different_densities}, for denser networks with \(M_0 = 4\) and \(M_0 = 5\), \emph{Heuristic-v3} achieves the highest robustness with average values of 0.412866 and 0.427933, respectively, demonstrating its effectiveness in denser networks. Similarly, in Table \ref{tab:table_results_real_world}, \emph{Heuristic-v3} achieves a robustness of 0.219386 on the EU Power Grid Network, outperforming the baseline algorithms.

\begin{minipage}[t]{0.5\textwidth}  
        \scriptsize
        \centering
        \captionof{table}{$R$ obtained on networks of different sizes.}
    \label{tab:table_results_different_nodes}
\setlength{\tabcolsep}{1mm}    
    \begin{tabular}{c|c|c|c|c}
    \hline 
        $N$ & Algs & Best & Worst & Average $\pm$ Variance \\
        \hline 
            & \emph{HC} & 0.276153 & 0.213245 & 0.251327 $\pm$ 0.000211 \\
            & \emph{SA} & 0.307163 & 0.221321 & 0.274463 $\pm$ 0.000269 \\
         100& \emph{SR} & 0.286163 & 0.211427 & 0.255401 $\pm$ 0.000229 \\
            & \emph{Heuristic-v1} & 0.274800 & 0.262700 & 0.270066 $\pm$ 0.000069 \\
            & \emph{Heuristic-v2} & \textbf{0.366399} & \textbf{0.355299} & \textbf{0.360699 $\pm$ 0.000020} \\
            & \emph{Heuristic-v3} & 0.324999 & 0.324599 & 0.324766 $\pm$ 0.000011 \\

        \hline 
            & \emph{HC} & 0.279601 & 0.214745 & 0.247127 $\pm$ 0.000141 \\
            & \emph{SA} & 0.281831 & 0.227268 & 0.257501 $\pm$ 0.000089 \\
         200& \emph{SR} & 0.277896 & 0.220101 & 0.253213 $\pm$ 0.000091 \\
            & \emph{Heuristic-v1} & 0.264925 & 0.254250 & 0.260791 $\pm$ 0.000058 \\
            & \emph{Heuristic-v2} & \textbf{0.354849} & \textbf{0.348724} & \textbf{0.351099 $\pm$ 0.000021} \\
            & \emph{Heuristic-v3} & 0.315075 & 0.303625 & 0.307550 $\pm$ 0.000028 \\
        \hline 
            & \emph{HC} & 0.264859 & 0.223717 & 0.243451 $\pm$ 0.000064 \\
            & \emph{SA} & 0.265958 & 0.227663 & 0.249812 $\pm$ 0.000091 \\
         300& \emph{SR} & 0.267519 & 0.230698 & 0.251128 $\pm$ 0.000063 \\
            & \emph{Heuristic-v1} & 0.249233 & 0.244899 & 0.247299 $\pm$ 0.000042 \\
            & \emph{Heuristic-v2} & \textbf{0.355855} & \textbf{0.348866} & \textbf{0.351699 $\pm$ 0.000027} \\
            & \emph{Heuristic-v3} & 0.309299 & 0.303022 & 0.307023 $\pm$ 0.000022 \\
            
        \hline 
            & \emph{HC} & 0.249601 & 0.222911 & 0.236645 $\pm$ 0.000038 \\
            & \emph{SA} & 0.249788 & 0.221655 & 0.238325 $\pm$ 0.000028 \\
        500 & \emph{SR} & 0.249093 & 0.228155 & 0.238423 $\pm$ 0.000033 \\
            & \emph{Heuristic-v1} & 0.233875 & 0.227803 & 0.230369 $\pm$ 0.000033 \\
            & \emph{Heuristic-v2} & \textbf{0.348228} & \textbf{0.344426} & \textbf{0.346460 $\pm$ 0.000024} \\
            & \emph{Heuristic-v3} & 0.305000 & 0.301996 & 0.303977 $\pm$ 0.000023 \\
   \hline          

    \end{tabular}
\end{minipage}
\hfill
\begin{minipage}[t]{0.5\textwidth}  

    \scriptsize
    \centering
    \captionof{table}{$R$ obtained on networks of varying edge densities.}
    \label{tab:table_results_different_densities}
\setlength{\tabcolsep}{1mm}    
    \begin{tabular}{c|c|c|c|c}
        \hline 
        $M_0$ & Algs & Best & Worst & Average $\pm$ Variance \\
        \hline 
          & \emph{HC} & 0.267699& 0.182465& 0.231189 $\pm$ 0.000272\\
          & \emph{SA} & 0.301401& 0.195479& 0.251081 $\pm$ 0.000367\\
        2 & \emph{SR} & 0.270889& 0.188913& 0.234173$\pm$ 0.000269\\
          & \emph{Heuristic-v1}& 0.266000& 0.260500& 0.262601 $\pm$ 0.000132\\
          & \emph{Heuristic-v2}& \textbf{0.346599}& \textbf{0.312231}& \textbf{0.333293 $\pm$ 0.000068}\\
          & \emph{Heuristic-v3}& 0.310200& 0.302000& 0.306233 $\pm$ 0.000088\\
        \hline 
          & \emph{HC} & 0.361012& 0.284971& 0.331501 $\pm$ 0.000214\\
          & \emph{SA} & 0.377793& 0.315786& 0.357099 $\pm$ 0.000132\\
        3 & \emph{SR} & 0.364866& 0.291878& 0.338351 $\pm$ 0.000182\\
          & \emph{Heuristic-v1}& 0.359299& 0.358399& 0.358866 $\pm$ 0.000041\\
          & \emph{Heuristic-v2}&  \textbf{0.402699}& \textbf{0.385199}& \textbf{0.394033 $\pm$ 0.000051}\\
          & \emph{Heuristic-v3}& 0.387299& 0.379999& 0.382566 $\pm$ 0.000063\\
        \hline 
          & \emph{HC} & 0.400000 & 0.365644 & 0.386593 $\pm$ 0.000059 \\
          & \emph{SA} & 0.413366 & 0.385940 & 0.401635 $\pm$ 0.000033 \\
        4 & \emph{SR} & 0.402673 & 0.370594 & 0.388876 $\pm$ 0.000030 \\
          & \emph{Heuristic-v1}& 0.400299& 0.395199& 0.397299 $\pm$ 0.000025\\
          & \emph{Heuristic-v2}& 0.308147& 0.302105& 0.305465 $\pm$ 0.000014\\
          & \emph{Heuristic-v3}& \textbf{0.416999}& \textbf{0.410799}& \textbf{0.412866 $\pm$ 0.000021}\\
        \hline 
          & \emph{HC} & 0.425451& 0.399378& 0.412409 $\pm$ 0.000024\\
          & \emph{SA} & \textbf{0.436733}& 0.410297 & 0.423445 $\pm$ 0.000017 \\
        5 & \emph{SR} & 0.421089 & 0.400198 & 0.412844 $\pm$ 0.000022 \\
          & \emph{Heuristic-v1}& 0.420999& 0.419599& 0.420266 $\pm$ 0.000005\\
          & \emph{Heuristic-v2}& 0.338199& 0.331899& 0.335066 $\pm$ 0.000004\\
          & \emph{Heuristic-v3}& 0.429299& \textbf{0.426099}& \textbf{0.427933 $\pm$ 0.000005}\\
        \hline 
    \end{tabular}

\end{minipage}
\hfill

\begin{table}[b]
\setlength{\tabcolsep}{1mm}
    \scriptsize
    \centering
    \caption{$R$ obtained on the EU Power Grid Network.}
    \label{tab:table_results_real_world}
\begin{center}

\begin{tabular}{c|c|c|c}
          \hline 
         Algs & Best & Worst & Average $\pm$ Variance \\  \hline 
         \emph{HC} &  0.213618&  0.210541& 0.212316 $\pm$ 0.000001\\  
             \emph{SA} &   0.215369&  0.195734& 0.206183 $\pm$ 0.000067\\
             \emph{SR} &   0.214041& 0.212116&  0.212764 $\pm$ 0.000001\\
             \emph{Heuristic-v1}& 0.211907& 0.195907& 0.205911 $\pm$ 0.000058\\
             \emph{Heuristic-v2}& \textbf{0.272040}& \textbf{0.270839}& \textbf{0.271440 $\pm$ 0.000002}\\
             \emph{Heuristic-v3}& 0.220805& 0.217966& 0.219386 $\pm$ 0.000002\\
            \hline 

    \end{tabular}
\end{center}
\end{table}

\section{Conclusion}

This paper proposes AutoRNet, an innovative framework designed to generate complete heuristics automatically which can improve the robustness of networks by integrating LLMs with EAs. AutoRNet uses NOS-based variation operations to create domain specific prompts for LLMs and an AFF to transfer hard constraint to soft one so that the searching space is relaxed and the searching process is more effective. The experimental results show that AutoRNet not only can design complete heuristics matching the complexity of manually ones by making use of advanced domain knowledge, but also can explore new strategies that yield better performance at the same practical cost benefiting from the AFF. Three best complete heuristics with different properties generated by AutoRNet were evaluated on both synthetic networks with varying sizes and densities and a real-world network, showing better performance over baseline algorithms. AutoRNet can significantly reduce the need for manual design and large datasets, providing a more flexible and adaptive solution. This leads to broader considerations about the potential of automated heuristic design in optimizing complex networks of the real world.



\bibliographystyle{elsarticle-num}


\newpage
\appendix
\section{The detailed population initialization prompt}

The population initialization is achieved by providing Large Language Models (LLMs) with detailed task specifications, which include a description of the problem, function signature, and functionality.
\begin{itemize}
    \item \textbf{Problem Description}: An explanation of the network robustness problem, detailing the optimization goals.
    \item \textbf{Function Signature}: Providing the function signature of the heuristic method to guide LLMs in generating the correct code.
    \item \textbf{Functionality}: Listing the specific functions and operations that the heuristic methods can directly call or use.
\end{itemize}

A typical task specification is shown in Figure A1:

Optionally, heuristic seed codes can be included in the prompt to guide the LLMs. It is already executable and serves as a template. The seed code used in the experiment is designed to modify a given network graph by randomly swapping edges and utilizes the \emph{compute\_robustness} function to evaluate the network's robustness after each modification, ensuring that only beneficial changes are kept, which is shown in Algorithm A1.\\

\begin{minipage}[c]{0.49\textwidth}
\centering
\includegraphics[width=1\columnwidth]{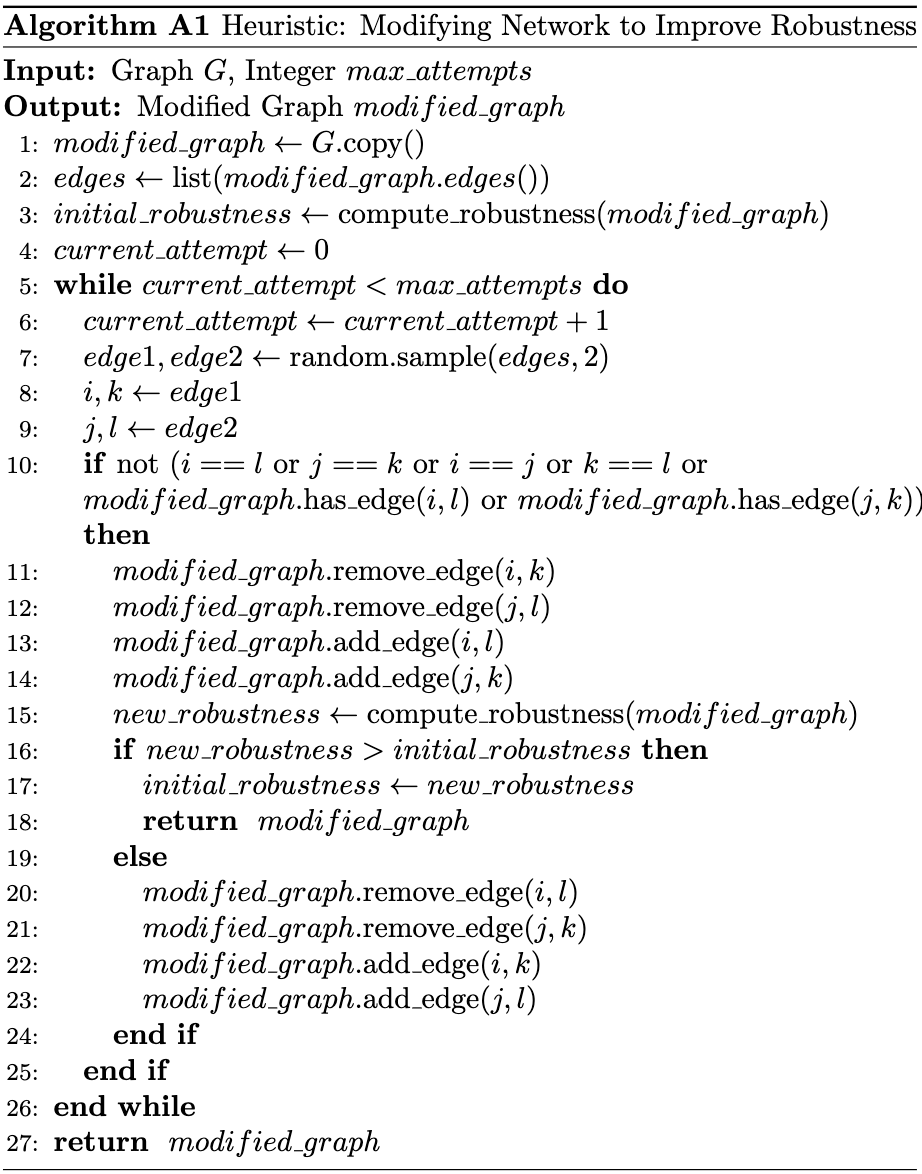}
    \end{minipage}
    \hfill
\begin{minipage}[c]{0.49\textwidth}
    \centering

\includegraphics[width=1\columnwidth]{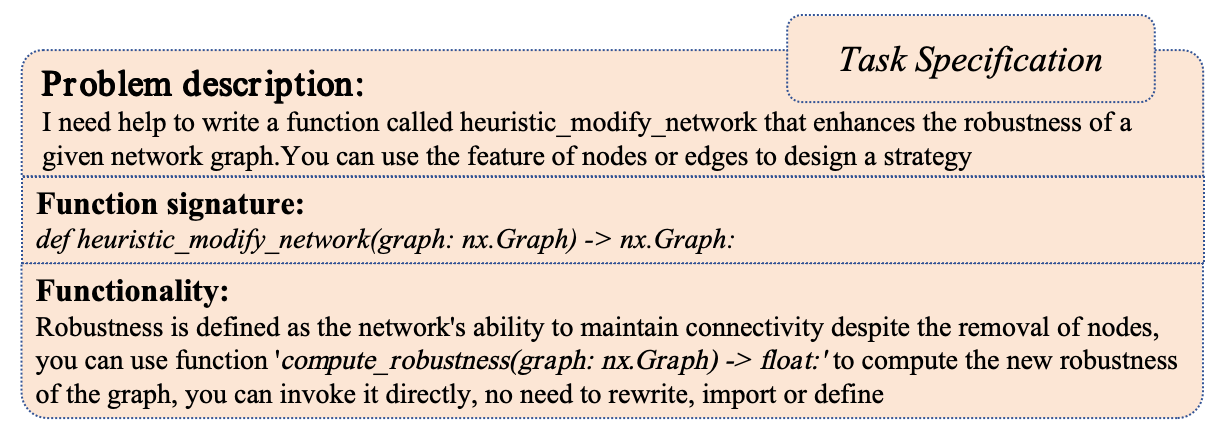}
\captionof{figure}{Task Specification for population initialization prompt}
\end{minipage}

\section{The detailed information of NOSs}
Network Optimization Strategies (NOSs) are a crucial component of AutoRNet. By integrating domain-specific knowledge into the variation operations. NOSs provide structured guidance that helps navigate the complex method space more effectively. 
Each NOS is composed of three main components: features, strategies, and actions. 

\begin{enumerate}
    \item \textbf{Features}: 
    \begin{itemize}
        \item \textbf{Degree}: The number of connections each node has.
        \item \textbf{Path Characteristics}: Shortest path, average path length, and network diameter.
        \item \textbf{Clustering Coefficient}: Local and global measures of how nodes tend to cluster together.
        \item \textbf{Connectivity}: Connected components and the strength of connections between them.
        \item \textbf{Centrality Measures}: Degree centrality, betweenness centrality, closeness centrality, and eigenvector centrality.
        \item \textbf{Edge Attributes}: Weight and direction of edges.
        \item \textbf{Dynamic Characteristics}: Robustness to failures and ability to recover.
        \item \textbf{Community Structure}: Tightly-knit groups within the network.
    \end{itemize}
    \item \textbf{Strategies}: 
    \begin{itemize}
        \item \textbf{High-Degree Node Priority}: Focus on nodes with many connections.
        \item \textbf{Low-Degree Node Priority}: Focus on nodes with fewer connections.
        \item \textbf{Betweenness Centrality Priority}: Focus on nodes that frequently appear on shortest paths.
        \item \textbf{Closeness Centrality Priority}: Focus on nodes that have short average distances to all other nodes.
        \item \textbf{Eigenvector Centrality Priority}: Focus on nodes that have high influence over the network.
        \item \textbf{High-Weight Edge Priority}: Focus on edges with higher weights.
        \item \textbf{Low-Weight Edge Priority}: Focus on edges with lower weights.
        \item \textbf{Shortest Path Optimization}: Optimize the shortest paths in the network.
        \item \textbf{Critical Path Optimization}: Optimize paths that are crucial for network performance.
        \item \textbf{Similarity-Based Node Selection}: Focus on nodes with similar attributes or roles.
        \item \textbf{Boundary Node Optimization}: Focus on nodes at the boundary of communities or clusters.
        \item \textbf{Homophily-Based Edge Optimization}: Focus on edges connecting nodes with similar attributes.
        \item \textbf{Heterophily-Based Edge Optimization}: Focus on edges connecting nodes with different attributes.
        \item \textbf{Hub-Peripheral Optimization}: Optimize the connectivity between hub nodes and peripheral nodes.
        \item \textbf{Random Node Selection}: Randomly select nodes for optimization to introduce variability.
        \item \textbf{Central Node Optimization}: Focus on nodes that are centrally located within their respective communities.
    \end{itemize}
    \item \textbf{Actions}:
    \begin{itemize}
        \item \textbf{Edge Addition}: Involves the addition of new edges to a network, thereby increasing its redundancy and robustness.
        \item \textbf{Edge Relocation}: Refers to the process of moving existing edges from one pair of nodes to another. This strategy alters the degree distribution of the nodes involved. 
        \item \textbf{Edge Swapping}: Involves exchanging the endpoints of two edges within the network. This technique preserves the original degree distribution.
    \end{itemize}
\end{enumerate}
The example of NOSs is shown in Figure A2:
\begin{figure}[t]
\centering
\includegraphics[width=1\columnwidth]{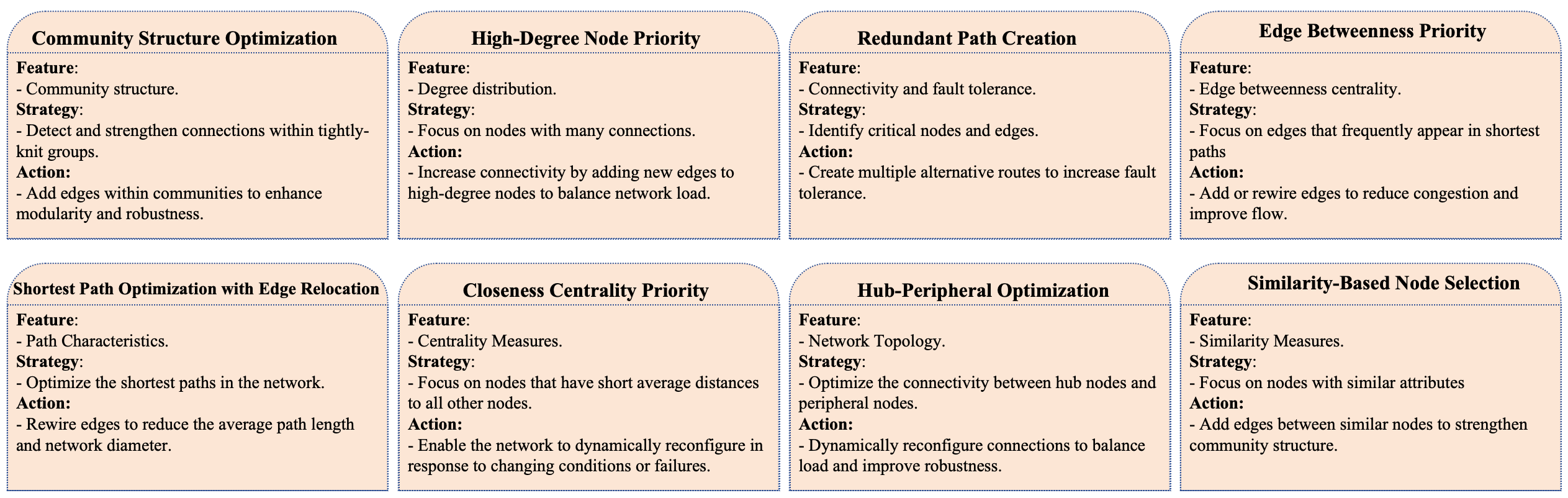}
\caption{Eight examples of NOSs. NOSs are randomly selected and included in the prompt for variation operation. }
\end{figure}
\section{C.The detailed Variation Operation prompt}
We define three types of Variation Operation prompts: \textbf{E1} generates offspring heuristics that are entirely different from the parent heuristics. \textbf{M1} modifies the current heuristic based on NOSs to enhance its effectiveness. \textbf{M2} fine-tunes the current heuristic to optimize its efficiency. Each type plays a specific role in the evolution process, balancing exploration and exploitation to enhance network robustness. M1 is given in Figure 3 of the main text, E1, M2 are given in Figures A3 and A4:

\begin{figure}[t]
\centering

\begin{minipage}[t]{\textwidth}
    \begin{minipage}[t]{0.45\textwidth}
    \centering
    \includegraphics[width=\textwidth]{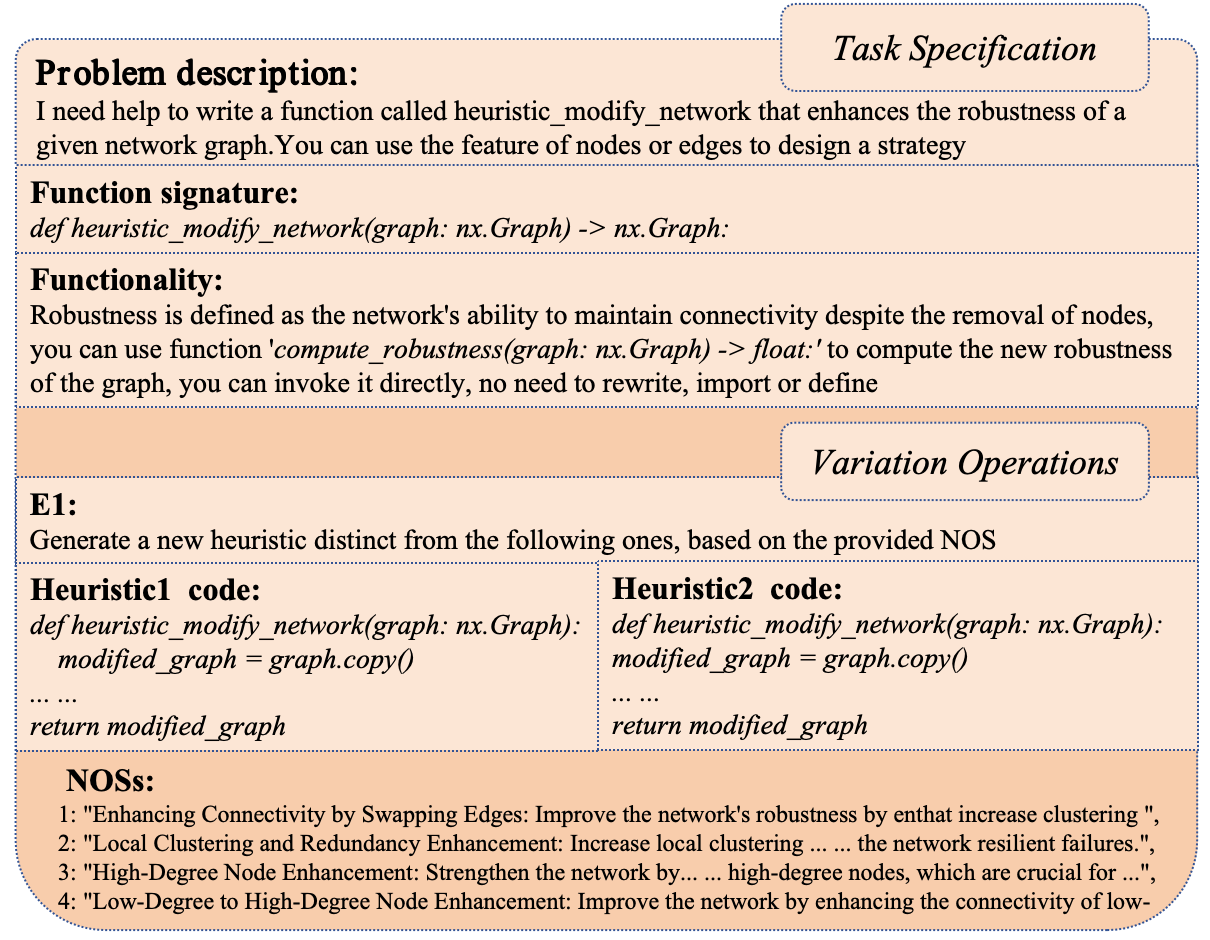}
    \caption{Variation Operation Prompt for E1. It helps to escape local optima by introducing new strategies and methods from NOSs.}
    \end{minipage}
    \hfill
    \begin{minipage}[t]{0.45\textwidth}
    \centering
\includegraphics[width=\textwidth]{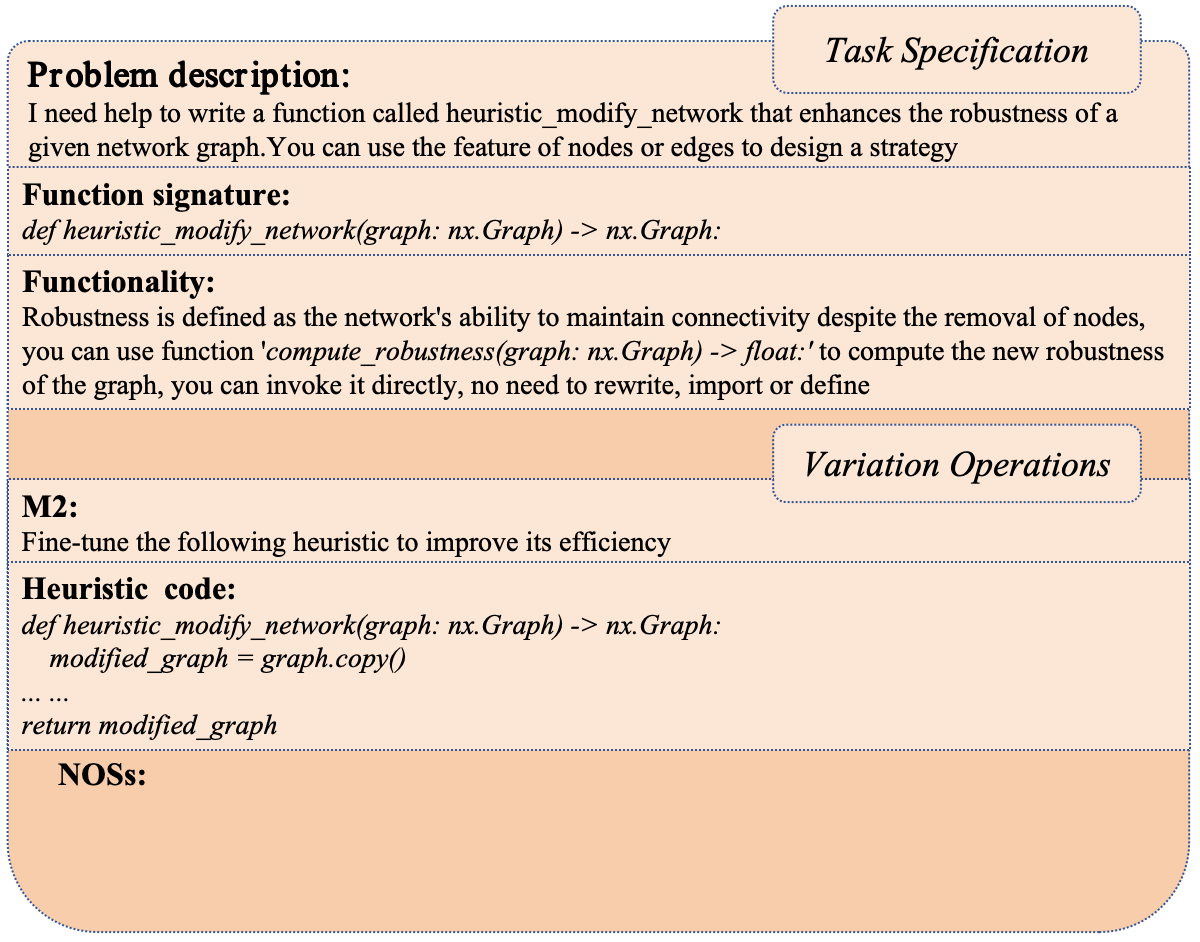}
\caption{Variation Operation Prompt for M2. It is a pure local search, making incremental adjustments to optimize the heuristic and don't have NOSs.}
    \end{minipage}
\end{minipage}

\end{figure}
\end{document}